# Adaptive Kalman-based hybrid car following strategy using TD3 and CACC

Yuqi Zheng[a], Ruidong Yan[a,*], Bin Jia[a,*], Rui Jiang[a], Adriana TAPUS[b]，Xiaojing Chen[c]，Shiteng Zheng[a], Ying Shang[a]

[a] Key Laboratory of Transport Industry of Big Data Application Technologies for Comprehensive Transport, School of Systems Science, Beijing Jiaotong University, 100044 Beijing, PR China

[b] Autonomous Systems and Robotics Lab, ENSTA Paris, Institut Polytechnique de Paris, 91120 PALAISEAU, France

[c] School of Economics and Management, Beihang University, 100044 Beijing, PR China

**ABSTRACT**

In autonomous driving, the hybrid strategy of deep reinforcement learning and cooperative adaptive cruise control (CACC) can fully utilize the advantages of the two algorithms and significantly improve the performance of car following. However, it is challenging for the traditional hybrid strategy based on fixed coefficients to adapt to mixed traffic flow scenarios, which may decrease the performance and even lead to accidents. To address the above problems, a hybrid car following strategy based on an adaptive Kalman Filter is proposed by regarding CACC and Twin Delayed Deep Deterministic Policy Gradient (TD3) algorithms. Different from traditional hybrid strategy based on fixed coefficients, the Kalman gain $H$, using as an adaptive coefficient, is derived from multi-timestep predictions and Monte Carlo Tree Search. At the end of study, simulation results with 4157745 timesteps indicate that, compared with the TD3 and HCFS algorithms, the proposed algorithm in this study can substantially enhance the safety of car following in mixed traffic flow without compromising the comfort and efficiency.

1. Introduction

Autonomous driving vehicles are expected to reduce traffic accidents, reduce energy consumption and enhance transportation efficiency (Chen et al., 2022; Grigorescu et al., 2020). Moreover, they have the potential to revolutionize the way individuals travel. (Kissai et al., 2019; Nikitas et al., 2021). Deep reinforcement learning (DRL) has demonstrated superior decision-making ability in complex competitive games such as Atari, Go chess, StarCraft, Dota 2, and Gran Turismo (Kaufmann et al., 2023) and has attracted extensive attention in the field of autonomous driving. It provides a new means for solving the decision-control challenges in autonomous driving and is considered the most promising control theory system for autonomous driving decision making problems.

Safety is a long-term existed challenge for autonomous driving (Muhammad et al., 2020). However, in DRL algorithms, the problems of inadequate training (Cao et al., 2021; Kalra et al., 2016) and weak generalization ability of neural networks (Liu et al., 2020; Chen[2] et al., 2021) are not well solved yet, and all of these factors may lead to the safety hazards to autonomous driving control strategies based on DRL. Therefore, it is urgent to improve the safety of autonomous driving based on DRL.

In the field of autonomous driving, there has been many researches about DRL algorithms that consider safety issues. Existing researches mainly focus on the following aspects: (1) the design of the reward function of DRL considering safety indicators; (2) the safety constraints on the exploration of DRL during the training process; and (3) the safety constraints on the output of DRL. And they are detailed below.

Firstly, the design of the reward function of DRL considering safety indicators. In reward function design, collisions are always recognized and a large negative value is set as the penalty (Pérez-Gil et al., 2022; Antonio and Maria-Dolores, 2023; Dong[1] et al., 2021; Ye et al., 2019), however, this reward function design approach ignores the state of the vehicle during a short period of time before the collision occurs. For this reason, the vehicle safety state during this small period of time can be described and incorporated into the safety reward function design by using the headway (Chen et al., 2020), time headway (Chen et al., 2023) and time to collision (TTC) (Selvaraj et al., 2023) or their correlation functions as indicators (He and Lv, 2023). In addition, a collision risk assessment function (Li et al., 2022) can be designed to use the risk assessment result as a reward value, thus to the risk-minimizing autopilot strategy.

Second, the safety constraints on the exploration of DRL during the training process. The constraints on the exploration are mainly external information guidance and risk avoidance (Hao et al., 2023). External information guidance includes various forms, such as expert guidance and rule-based constraints: expert guidance uses expert knowledge to establish a safe state-action space, restricting behavioral exploration to within the safe state-action space (Hoel et al., 2019). Rule-based constraints constrain the exploration of the agent through certain rules or strategies to improve the safety of exploration (Bouton et al., 2019; He et al., 2023). Risk avoidance, on the other hand, recognizes collision risks and guides agents to avoid dangerous actions which may lead to unsafe state (Mukadam et al., 2017).

Third, the safety constraints on the output of DRL. Constraining on the outputs of DRL can effectively improve the policy safety. In hazardous conditions, actions from rule-based policies will replace the RL actions to improve safety, such as emergency brakes (Hwang et al., 2022), empirically based safety actions (Baheri et al., 2020), and actions generated by MPC algorithms (Bautista-Montesano et al., 2022). Besides, in the case of poor performance of RL strategies, the action outputs will be constrained to improve safety while improving the overall performance (Yang et al., 2023; Cao et al., 2022).

Among the three types of methods, the third method directly imposes safety constraints on the action outputs, which can maximize the safety of autonomous driving, and has attracted much attention in recent years. Yan et al. (2022) utilizes cooperative adaptive cruise control (CACC) to constrain the output of DRL. It calculates the action and reward generated by CACC and deep deterministic policy gradient respectively at each timestep, and mixes the action outputs with the corresponding fixed coefficients by comparing the reward values. However, in mixed traffic flow scenarios, the hybrid strategy based on fixed coefficient is difficult to adapt to the complex environments, which may lead to the degradation of the following performance and even unsafe accidents.

Regarding to the above problems, this paper proposes a hybrid car following strategy based on an adaptive Kalman Filter (KF). The strategy regards CACC and Twin Delayed Deterministic Policy Gradient (TD3) as two parallel control signals, fuses these two signals using KF and realizes the adaptive adjustment of the coefficients of KF. This paper has the following three main contributions:

A. Different from the fixe-coefficient strategy with single step prediction (Yan et al., 2022), this paper proposes a hybrid car following strategy based on KF, which improves from fixed coefficients to adaptive coefficients by performing multiple-step predictions, and thus significantly improves the car following performance of autonomous vehicles under mixed traffic flow scenarios.

B. Unlike the traditional KF algorithm, at each timestep, the prediction error covariance $P$ is computed iteratively using a multi-step prediction method, the measurement noise covariance $\mathcal{R}$ is computed using a MCTS algorithm, and the Kalman gain $H$ is adaptively adjusted according to $P$ and $\mathcal{R}$.

C. Through the simulation under mixed traffic flow scenarios with 4157745 timesteps, the algorithm proposed in this paper successfully reduces the number of collisions from 58 (TD3) times and 1 (Hybrid Car Following Strategy, HCFS) to 0 (Adaptive Kalman Hybrid Car Following Strategy, AK-HCFS) without reducing comfort and efficiency. And it significantly improves the safety of car following of autonomous vehicles under mixed traffic flow scenarios.

The rest of this paper will be organized as follows: Section 2 reviews the existing research HCFS, Section 3 describes the algorithm proposed in this paper, Section 4 shows the detail of the simulation, and Section 5 draws conclusions.

**2. Review of Conventional Hybrid Car Following Strategy**

Yan et al. (2022) has proposed a car following algorithm for autonomous driving based on the Deep Deterministic Policy Gradient algorithm. This study employs CACC algorithm as a guardian and combines it with the deep reinforcement learning algorithm. This algorithm successfully enhances the performance of car following. In this paper, the Deep Deterministic

Policy Gradient algorithm is replaced by the TD3 algorithm. The Equation (1) calculates the acceleration $a_k$ at timestep $k$. In this equation, $\alpha$ and $\beta$ are coefficients, $a_{TD3\_k}$ represents the acceleration obtained by the TD3 algorithm at timestep $k$, and $a_{CACC\_k}$ represents the acceleration obtained by the CACC algorithm at timestep $k$. When the reward of TD3 is higher than that of CACC, where TD3 algorithm is used alone, namely both $\alpha$ and $\beta$ are set to 0. Otherwise, $\alpha$ is set to 1, and $\beta$ is set to 0.5.

$$a_k = \begin{cases} (1-\beta)a_{TD3\_k} + \beta a_{CACC\_k}, & r_{TD3\_k} > r_{CACC\_k} \\ (1-\beta)a_{CACC\_k} + \beta a_{TD3\_k}, & \text{else} \end{cases} \quad (1)$$

$$\beta = \begin{cases} 0, & \alpha=0 \\ 0.5, & \alpha=1 \end{cases} \quad (2)$$

$$a_{CACC\_k} = (v_{ego\_k+1} - v_{ego\_k})/\Delta T \quad (3)$$

$$v_{ego\_k} = v_{ego\_k-1} + k_p e_k + k_d \dot{e}_k \quad (4)$$

$$e_k = x_{front\_k} - x_{ego\_k} - L - t_{hw} v_{ego\_k} \quad (5)$$

The CACC algorithm adopted in this study is generally recognized and it has been verified through on-road tests (Milanés et al., 2013; Milanés and Shladover, 2014). The model is shown in (3) ~ (5), (3) calculates the acceleration where $\Delta T$ represents the time interval. The speed of ego vehicle at timestep $k$ is calculated in (4). where $k_p$ (0.45) and $k_d$ (0.25) are validated parameters. $e_k$ is the difference between the expected distance and the real distance, which is calculated by (5). $x_{front\_k}$ and $x_{ego\_k}$ are the position of the preceding vehicle and the position of the ego vehicle respectively. $L$ is the length of the vehicle, and $t_{hw}$ represents the time headway, which is set as 0.6 s in the study of Milanés and Shladover (2014).

The pseudo-code of this algorithm is given in Pseudo-code 1. At each timestep, the reward of the actions obtained by CACC and TD3 algorithms will be calculated. Then, a combination will take place if the reward of TD3 is not higher than that of CACC.

| | Pseudo-code 1 |
|---|---|
| 1 | **for** $k = 0$ to Num **do** |
| 2 |    Observe initial state: |
| 3 |      Choose accelerations $a_k$ with exploration noise $\varepsilon$ |
| 4 |      Calculate $s'$, $r_{CACC}$, $r_{TD3}$ |
| 5 |     **if** $\alpha = 0$: |
| 6 |       $\beta = 0$ |
| 7 |     **if** $\alpha = 1$:<br>      $\beta = 0.5$ |
| 8 |     **if**: $r_{CACC} > r_{TD3}$ |
| 9 |       $a_k = (1-\beta)a_{TD3\_k} + \beta a_{CACC\_k}$ |
| 10 |     **else**: |
| 11 |       $a_k = (1-\beta)a_{CACC\_k} + \beta a_{TD3\_k}$ |
| 12 |    Sample random mini-batch of $N$ transitions from $D$ |
| 13 |    Set $y \leftarrow r + \gamma \min_{i=1,2} Q_{\theta_i'}(s',a)$ |
| 14 |    Update critic through minimizing loss: $\sum (y - Q_i(s,a))^2$ |
| 15 |    Update actor policy using deterministic policy gradient: |
| 16 |       $\nabla_\phi \ J(\phi) = N^{-1} \sum \nabla_a Q_{\theta 1}(s,a) \big|_{a=\pi_\phi(s)} \nabla_\phi \pi_\phi(s)$ |
| 17 |    Update target networks with soft update |
| 18 | **end for** |

***Remark 1***: The HCFS algorithm combines the CACC algorithm and the car following algorithm based on the deep reinforcement learning algorithm. However, when giving the hybrid strategy, 1. The predictions are made only for one step and take no account for the potential hazards. 2. The selection of fixed coefficients does not account for the changes of state at each timestep. In a real-world driving environment, autonomous vehicles will be affected by a variety of factors (Melman et al., 2021), and the fixed coefficients $\beta$ may not adapt to the complex environments well. Therefore, in a complex traffic environment, the use of fixed coefficients under the single-step prediction may lead to unstable performance of autonomous vehicles, and even result in accidents and safety hazards.

### 3. Adaptive Kalman-based Hybrid Car-Following Strategy

To address the problems in Remark 1, this study proposes an adaptive Kalman-based hybrid car following strategy using TD3 and CACC. In section 3.1, a Markov decision process (MDP) model for car following scenario of multi-vehicles is constructed. Then, Section 3.2 gives the detail of the proposed algorithm.

**3.1 Markov decision process (MDP)**

A MDP model is tailored to the car following problem with the scenario of multi-vehicles on highway. A MDP model consists of a state set, an action set and a reward function (Sutton and Barto, 1998). In this study, the state set consists of the distance difference between the preceding vehicle and the following vehicle $x_{error\_k}$, the speed difference between the preceding vehicle and the following vehicle $v_{error\_k}$, the velocity of the ego vehicle $v_{ego\_k}$, the distance difference between the leading vehicle and the ego vehicle $x_{error\_0\_k}$ and the speed difference between the leading vehicle and the ego vehicle $v_{error\_0\_k}$. The expression of state is shown in Equation (6).

$$\{x_{error\_k}, v_{error\_k}, v_{ego\_k}, x_{error\_0\_k}, v_{error\_0\_k}\} \quad (6)$$

The action set is the set of the longitudinal acceleration of the autonomous vehicle, and the expression is shown in Equation (7).

$$\{a_k\}, a_k \in [-a_{bound}, a_{bound}] \quad (7)$$

In autonomous driving, safety, comfort, and efficiency are the most common composition of the reward function. In addition, the stability should also be taken into consideration, since it is a multi-vehicle scenario. Therefore, the reward function consists of four parts, namely, stability, comfort, safety and efficiency.

(a) Reward of stability

The speed difference between the vehicles in the platooning ought to be kept as small as possible to maintain the stability. Therefore, the speed difference between the leading vehicle and the ego vehicle is considered in the reward function (Yan et al., 2022) The expression of the reward of stability $f_{stab\_k}$ is shown in (8),

$$f_{stab\_k} = -\frac{|v_{error\_0\_k}|}{v_{max}} \tag{8}$$

where $v_{max}$ is the maximum speed

(b) Reward of comfort

The comfort reward $f_{cft}$ is related with the change rate of acceleration $Jerk_k$ (Yan et al., 2022) and its' equation is shown in Equation (9). The $a_{bound}$ is the boundary value of acceleration and in this study, it is usually set as 3. Equation (10) calculates $Jerk_k$. The closer to 0, the more comfortable it is.

$$f_{cft} = -\frac{|Jerk_k|}{2a_{bound}\Delta T} \tag{9}$$

$$Jerk_k = (a_{perform\_k} - a_{perform\_k-1})/\Delta T \tag{10}$$

(c) Reward of safety

To improve the safety performance, safety reward is taken into consideration in the reward function. It is calculated in Equation (11). Time left to collision $TTC_k$ represents the time left to collision. It calculates the remaining time for the collision between the preceding vehicle and the ego vehicle and its' equation is given by Equation (12). When the $TTC_k$ is between 0 and 2.7 seconds (Zhu et al., 2020) the safety reward will be a negative value as a punishment.

$$f_{safe\_k} = \begin{cases} \log(\dfrac{TTC_k}{2.7}), & 0 \leq TTC_k \leq 2.7 \\ 0, & \text{otherwise} \end{cases} \quad (11)$$

$$TTC_k = -\dfrac{x_{error\_k}}{v_{error\_k}} \quad (12)$$

(d) Reward of efficiency

As to efficiency, this study sets an optimal time headway ($T_{desire}$, 0.6 seconds) for the following vehicles. All the following vehicles are expected to keep the optimal time headway and any deviation form it will be punished. The efficiency reward $f_{eff\_k}$ is calculated in (13). $x_{exp\_k}$ is the expected distance and it is given in (14). $x_k$ represents the actual distance while $x_{max}$ represents the maximum distance.

$$f_{eff\_k} = -\dfrac{|x_{exp\_k} - x_k|}{x_{max}} \quad (13)$$

$$x_{exp\_k} = v_{ego\_k} \cdot T_{desire} \quad (14)$$

After all, the reward function at timestep $k$ is shown in Equation (15).

$$\text{reward}_k = f_{stab\_k} + f_{cft\_k} + f_{safe\_k} + f_{eff\_k} \quad (15)$$

**3.2 The proposed algorithm**

The structure of the AK-HCFS proposed by this study is shown in Fig. 1 a1 and HCFC proposed by Yan et al. (2022) is shown in Fig. 1 a2 are all shown in Fig. 1. Different from HCFS, the proposed algorithm makes more prediction steps. More importantly, the proposed algorithm uses adaptive KF to get the adaptive parameter $H$. Besides, the measured noise covariance $R$ of KF is calculated by MCTS and the predicted error covariance $P$ is iteratively obtained by KF.

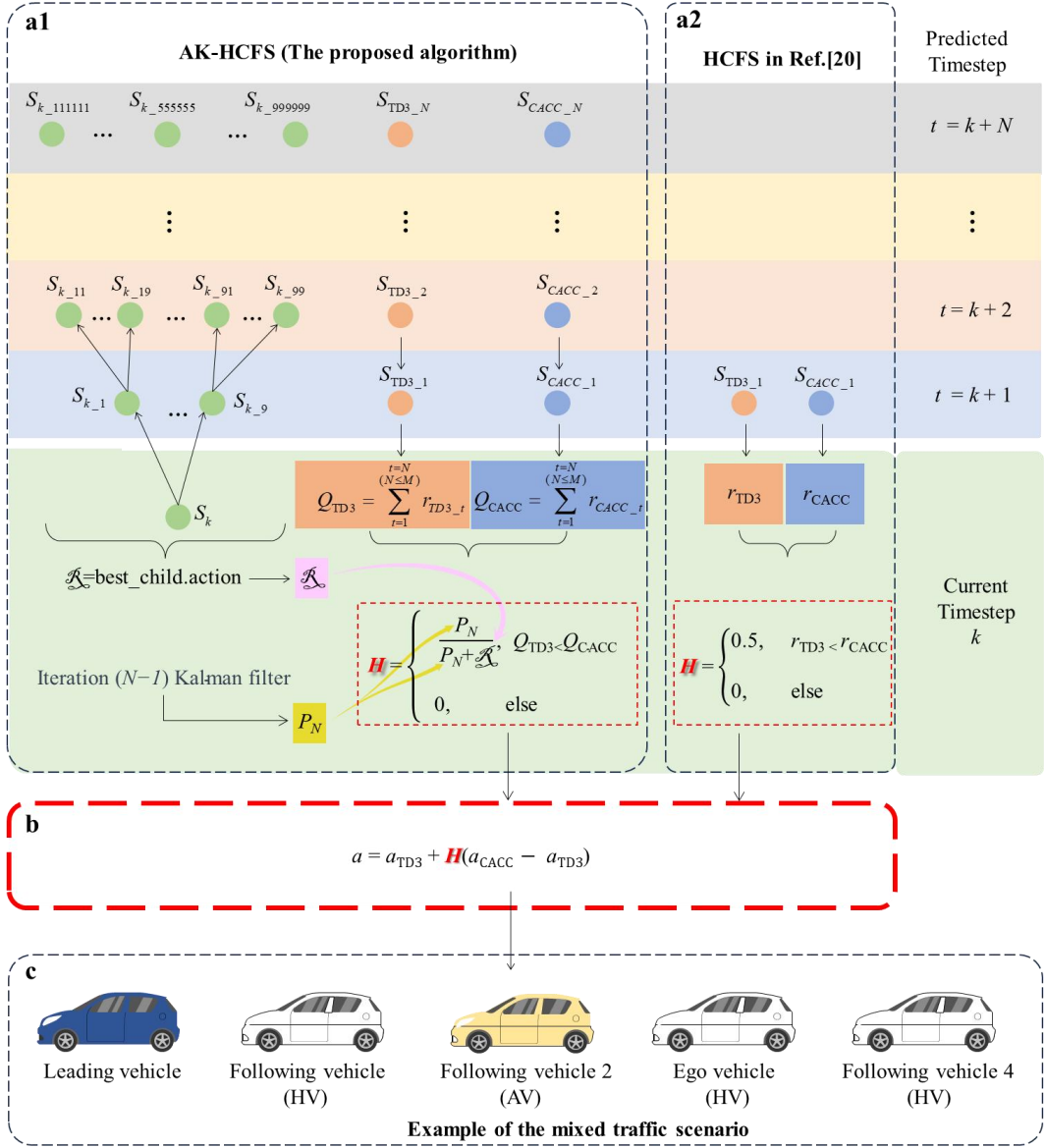

Fig. 1. Structure of the proposed algorithm. **a1**, The structure of the AK-HCFS. **a2**, The structure of HCFS. It only cares about the next state at timestep $t=k+1$ and it compares the reward of the TD3 and CACC. In case the reward of TD3 is lower than that of CACC, the fixed coefficient is chosen as 0.5. **b**, The calculation of the acceleration command using KF in terms of parameter $H$. **c**, An example of the mixed traffic scenario. The blue vehicle is the leading vehicle and others are all following vehicles. The yellow vehicle is the autonomous vehicle controlled by the acceleration command of step b. The white vehicles are human drive vehicles.

The procedure of the proposed algorithm is shown in Fig. 1 (a1). First, it will predict the future states and get actions from CACC and TD3 respectively at timestep $t=k+N$. Then the adaptive coefficients $H$ of KF is calculated by $\mathcal{R}$ and $P_N$, where the KF will iterate $N-1$ times to get $P_N$ and the

MCTS is applied to get the $\mathcal{R}$. In the end, the acceleration is calculated with $H$. And the calculation of the acceleration command is given by:

$$a = a_{TD3} + H(a_{CACC} - a_{TD3}) \quad (16)$$

$$H = \begin{cases} \dfrac{P_N}{P_N + \mathcal{R}}, & \sum_{t=1}^{t=N(N \leq M)} \gamma^{t-1} r_{TD3\_t} < \sum_{t=1}^{t=N(N \leq M)} \gamma^{t-1} r_{CACC\_t} \\ 0, & \text{else} \end{cases} \quad (17)$$

where $a$ is the acceleration command calculated by Equation (16), $a_{TD3}$ is the output of TD3 algorithm and $a_{CACC}$ is the output of CACC algorithm. $H$ is the adaptive Kalman parameter calculated by Equation (17), $\gamma$ is the depreciation coefficient, $P_N$ is the prediction error covariance of the KF, and $\mathcal{R}$ is the measurement noise covariance of the KF.

As can be seen form (16), the adaptive Kalman parameter $H$ is the key of the proposed algorithm. Moreover, the value of $H$ is calculated by Equation (17) via variables $P_N$, $\mathcal{R}$, and cumulative rewards of TD3 and CACC. The value of $P_N$ in terms of KF will be given in Part A of this section. The value of $\mathcal{R}$ is given by MCTS, and the specific calculation is given in Part B of this section. The cumulative rewards of TD3 and CACC are calculated through the prediction part as shown in Fig. 1 (a1), and Pseudo-code 2 gives the pseudo-code of the prediction part.

**Pseudo-code 2**

| | |
|---|---|
| 1 | Observe state at timestep k: |
| 2 |     Get accelerations $a_{TD3\_k}$ from TD3 |
| 3 |     Set $t$=1, $H$=1 $R_{CACC} = 0$, $R_{TD3} = 0$, $a = a_{TD3\_k}$ |
| 4 |     **while** $n < M$: |
| 5 |         $s_{CACC\_t}, s_{TD3\_t} = s_k$ |
| 6 |         Calculate $a_{CACC\_t}$, $r_{CACC\_t}$, $s'_{CACC\_t}$ and $a_{TD3\_t}$, $r_{TD3\_t}$, $s'_{TD3\_t}$ |
| 7 |         $R_{CACC} = R_{CACC} + \gamma^{n-1} r_{CACC}$, $R_{TD3} = R_{TD3} + \gamma^{n-1} r_{TD3}$ |
| 8 |         **if**: $R_{CACC} > R_{TD3}$ : |
| 9 |             $N=t$ |
| 10 |             Get $H$ from KF |
| 11 |             **break** |

12      $a = a_{TD3} + H(a_{CACC} - a_{TD3})$

***Remark 2***: Different from the fixe-coefficient approach based on single-step prediction in the literature (Yan et al., 2022), this paper proposes a hybrid car following strategy based on adaptive KF for multiple-step predictions. The prediction error covariance $P$ is computed iteratively using the multi-step prediction method, and the measurement noise covariance $\mathcal{R}$ is computed using the MCTS algorithm, and the Kalman gain $H$ is adaptively adjusted according to $P$ and $\mathcal{R}$. The fixed coefficients in literature (Yan et al., 2022) are improved to adaptive coefficients, which improves the ability to adapt to the complex environments, and thus significantly improves the car following performance of autonomous vehicles under mixed traffic flow scenarios.

**A) Value of $P_N$ in terms of KF**

KF has been widely used in industry. The selection of parameters of KF is particularly important in practical problems and many studies focused on this issue. For example, Zhang (2018) updates the parameters by minimizing the variance of the output; Fu and Cheng (2022) have updated the parameters by maximizing the posterior probability distribution of the system state. There are also some studies combining machine learning with KF (Yean et al., 2017; Chen[1] et al., 2021). However, these methods aim to make the estimated values closer to the actual values and are not suitable for the hybrid problems.

Different from above researches, the KF updates the predicted error covariance $P_N$ by iteration and updates measurement noise covariance $\mathcal{R}$ by MCTS (see Part B of this section). The model of the iteration is shown in Equations (18) ~ (20).

$$P_{kal\_n} = A_{kal\_n} + Q_{measure} \tag{18}$$

$$K_{kal\_n} = \frac{P_{kal\_n}}{P_{kal\_n} + \mathcal{R}_{measure}} \tag{19}$$

$$A_{kal\_n+1} = (1 - K_{kal\_n})P_{kal\_n} \tag{20}$$

where, $P_{kal\_n}$ is the predicted error covariance at the step *kal_n* of KF. $A_{kal\_n}$ represents the accumulated error and $Q_{measure}$ represents process noise covariance. Equation (19) calculates the Kalman gain at the *kal_n* step of the KF, and Equation (20) updates the cumulative error.

The main steps of the KF are: First, initialize the KF. $Q_{measure}$ is set as 0.01, the initial measurement noise covariance $\mathcal{R}_{measure}$ is set as 0.01 and the initial cumulative error $A_1$ is set as 1. Subsequently, the KF algorithm is iterated $N-1$ times and $P_N$ is the last. $P_{kal\_n}$. Then, the final measurement noise covariance $\mathcal{R}$ is obtained from MCTS. Finally, the adaptive parameter $H$ is calculated. The Pseudo-code 3 is shown below.

---

**Pseudo-code 3**

1. Initialize KF, $Q_{measure}$ =0.01, $\mathcal{R}_{measure}$ =0.01, A=1, get $N$, *kal_n*=1.
2.     **while** *kal_n* < $N$-1 :
3.         $P_{kal\_n} = A_{kal\_n} + Q_{measure}$
4.         $K_{kal\_n} = \dfrac{P_{kal\_n}}{P_{kal\_n} + \mathcal{R}_{measure}}$
5.         $A_{kal\_n+1} = (1 - K_{kal\_n})P_{kal\_n}$
6.         *kal_n*=*kal_n*+1
7. $P_N = P_{kal\_N}$
8. Get $\mathcal{R}$ from MCTS
9. $H = \begin{cases} \dfrac{P_N}{P_N + \mathcal{R}}, & \sum_{t=1}^{t=N(N \leq M)} \gamma^{t-1} r_{TD3} < \sum_{t=1}^{t=N(N \leq M)} \gamma^{t-1} r_{CACC} \\ 0, & \text{else} \end{cases}$

---

**B)** Value of $\mathcal{R}$ in terms of **MCTS**

The key steps of the MCTS algorithm include selection, expansion, simulation, and back propagation. Among them, selection means selecting child nodes through a certain strategy. The

Upper Confidence Bounds for Trees (UCB1) formula used in this study is shown in Equation (21),

$$UCB1 = \frac{\sum \gamma^{l_{node}-1} Q_i}{visit\_count} + C\sqrt{\frac{\ln(visit\_count_p)}{visit\_count}} \qquad (21)$$

where $\sum \gamma^{l_{node}-1} Q$ is the cumulative reward of the node. C is the exploration constant, and it is set as 7. $visit\_count_p$ represents the number of visits to the parent node.

Expansion means selecting unexplored child nodes for expansion when the selected node is not fully expanded. Simulation means starting from the expanded node and executing the possible actions. In this study, the simulation process is the process of taking an action as $\mathcal{R}$ and applying it to obtain H. The reward function is consistent with Equation (10). Back propagation means that after the simulation ends, the simulation results are propagated back to the selected nodes, and their visit counts and cumulative rewards are updated.

Additionally, the iteration is set to 1000 times. During the calculation, the expansion will be terminated when a child node collides or the child node reaches the maximum step of prediction. The output of this game is the value of $\mathcal{R}$ which is the action of the root node's child with the most visit counts. Pseudo-code 4 for the MCST algorithm is shown below.

| **Pseudo-code 4.** |
|---|
| 1    Get state, $N$ |
| 2    **for** _ in range(number of explorations): |
| 3        node=root node |
| 4        **while** node has children: |
| 5           **if** random.random() > 0.1: |
| 6              node = max(node.children, key=lambda child:ucb1(child, exploration_constant)) |
| 7              C=C×0.995 |
| 8           **else**: |
| 9              node = random.choice(node.children) |
| 10      **while** node has no.children and node is not terminal: |
| 11          apply all the possible actions |
| 12          generate new node |
| 13      r = node.reward |

| 14 | **while** node: |
| 15 |     node.visit_count += 1 |
| 16 |     node.action_value += $\gamma^{l_{node}-1} r$ |
| 17 |     node = node.parent |
| 18 | best_action = max(root_node.children, key=lambda child: child.visit_count).action |

## 4. Simulation

To verify the effectiveness of the proposed algorithms, we compare and analyze the three algorithms of TD3, HCFS, and AK-HCFS. The followings will introduce from the three aspects: environment and data, training and setting, testing result and analysis.

### 4.1 Environment and data

The data NGSIM US101 (Alexiadis et al., 2004) is used in this study, which was collected from the southbound highway in Los Angeles, California, USA. The diagram of the road section is shown in Fig. 2. First, To avoid the impact of merge and lane change, we selects car following data from Lane 1-4 and deletes the data involving lane changes. Then, the data less than 20 seconds is removed. Finally, the method from Dong[2] et al. (2021) is used to reconstruct the data. There are 1520 car following events in total and they are divided into the train set (1064 events) and test set (456 events) randomly.

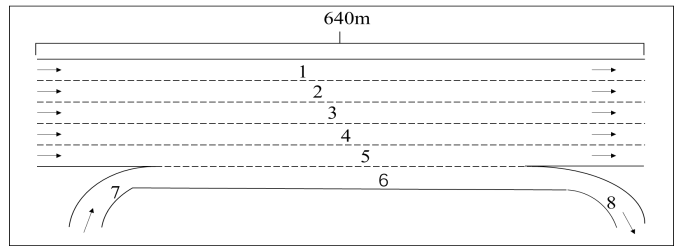

Fig. 2. Road segment of the dataset. The section is about 640 meters and has 8 lanes, in which lane 7 is the enter ramp and lane 8 is the exit ramp. Lane 6 is connected with lane 7 and 8, so it will be influenced by the insert vehicles, as a result the data from lane 6, 7, and 8 are all discard.

This study builds a multi-vehicle car following environment for simulation. The speed and position data of the leading vehicle are extracted from the NGSIM data set. The following vehicles would be either autonomous vehicles or human drive vehicles. And autonomous driving vehicles are controlled by reinforcement learning algorithms. Human-drive vehicles are controlled by Intelligent Driver Model (IDM, Treiber and Kesting (2013))).

The initial time headway of the preceding and following vehicles is set as 0.6 seconds, and the initial speed of following vehicles is consistent with that of the leading vehicle. In addition, a vehicle kinematic Equation (22) is used in this paper,

$$a_{cmd} = a + \tau \dot{a} \tag{22}$$

where $a_{cmd}$ is the control command, $a$ is the upper-level command, and $\tau$ is the time delay (0.4 seconds).

The IDM proposed by Treiber and Kesting (2013) is applied to control the human drive vehicles. The IDM this study used is shown in Equations (23) ~ (24). Equation (23) calculates the change rate of speed at timestep $k$, where $a_{max}$ is the maximum acceleration, $v_{desire}$ is the desired speed and $d^*$ is the desired distance, which is calculated by Equation (24),

$$\dot{v}_k = a_{max} \left[ 1 - (\frac{v_{k-1}}{v_{desire}})^4 - (\frac{d^*(v_{k-1}, \Delta v_{k-1})}{d})^2 \right] \tag{23}$$

$$d^*(v_k, \Delta v_k) = d_0 + Tv_k + \frac{v_k \Delta v_k}{2\sqrt{a_{max} b}} \tag{24}$$

where $d_0$ represents the minimum distance, $T$ represents the time headway, $\Delta v_k$ represents the speed difference between the preceding and following vehicles and $b$ is the comfortable deceleration.

The parameters of the IDM model are as shown in Table 1.

Table 1. Parameters of IDM

| Parameter | Value |
| --- | --- |
| $a_{max}$ | 3.79 |
| $d_0$ | 1.08 |
| $b$ | 3.5 |
| $v_{desire}$ | 39.48 |
| $T$ | 1.22 |

**4.2 Training and setting**

In the training environment, the speed and position of the leading vehicles are all from the train set (1064 events). There are 3 following vehicles, mixed with autonomous vehicles and human drive vehicles. All the human drive vehicles are controlled by the IDM, and autonomous vehicles are controlled by TD3, HCFS, and AK-HCFS respectively.

The TD3 algorithm proposed by Fujimoto et al. (2018) is used in this simulation. It is mainly composed of two sets of Actor-Critic networks and an experience buffer. The actor networks are composed of an input layer, an output layer, and two hidden layers of 32 neurons and 16 neurons respectively. The activation function of the hidden layer is the Tanh function. The setting of parameters of the TD3 algorithm are shown in Table 2.

Table 2 Settings of the parameters

| Parameters | Value |
| --- | --- |
| Memory_size | 20000 |
| Learning rate | 0.001 |
| policy_delay | 500 |
| act_noise | 0.1 |
| target_noise | 0.2 |

**4.3 Result and analysis**

In the testing environment, the speed and position of the leading vehicles are all from the test set (456 events) randomly. To examine the generalization performance of these three algorithms in complex environments, this paper uses a four-vehicle following scenario for testing. And all the human-drive vehicles are controlled by IDM, and autonomous vehicles are controlled by TD3, HCFS, and AK-HCFS respectively.

The analysis is conducted on both the statistical results and individual cases. First, this paper analyzes the statistical results of the tests for all scenarios, and the overall performance of the three algorithms can be obtained. Second, this paper analyzes the case where HCFS has the collision and a classical case where no collisions occurred.

A. **Statistical result**

We have simulated all the car following events in the test set and gone through all the possible scenarios for the following vehicles (a total of 6,840 car following events). Four indicators, including collisions in the tests, the mean value of *TTC*, the absolute value of jerk, and the average speed, are statistically analyzed.

(1) Statistical analysis of collisions in the tests

The statistical results show that the TD3 algorithm has 58 collisions, while the HCFS algorithm has 1 collision, and the proposed algorithm has no collisions. This shows that the proposed algorithm largely improves the safety of car following of autonomous vehicles. It is important to note that all the collisions are from the fourth vehicle, and there is no collision in 3-following-vehicle scenarios. It indicates that the proposed algorithm is more adaptable in complex environments, and it is safer than the other two algorithms.

Table 3. Collisions of all the tests

| Algorithm  | TD3 | HCFS | K-HCFS |
|------------|-----|------|--------|
| collisions | 58  | 1    | 0      |

(2) Statistical analysis of the mean value of *TTC*

Figs. 3, 4 and 5 show the quantitative statistical results of following vehicles of which the mean value of *TTC* is between 0~2.7 seconds, the numerical statistical results and the quantitative statistics of *TTC* between 0~2.7 under different leading vehicles of different speed respectively. Table 4 shows the minimum, maximum, median, and the first quartile Q1 and the third quartile Q3 in Fig. 4.

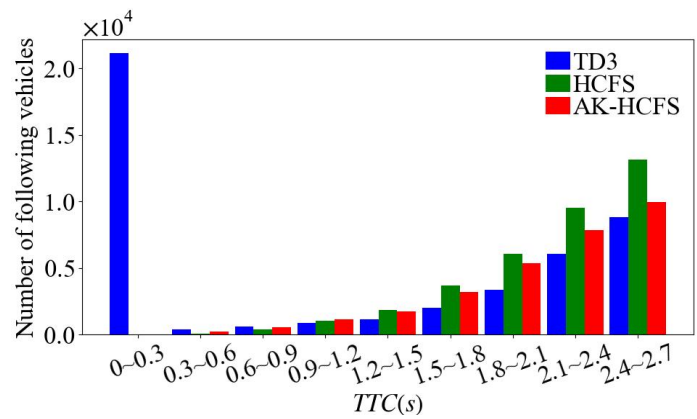

Fig. 3. Statistics of the number of following vehicles of which *TTC* within the range of 0 to 2.7 seconds of the three algorithms. The horizontal axis represents different intervals of *TTC* and the vertical axis represents the amount of the following vehicles.

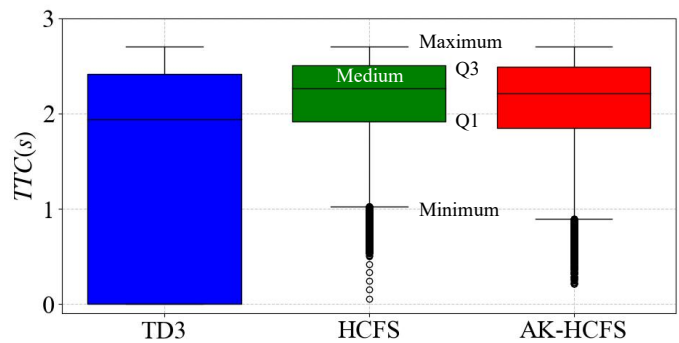

Fig. 4. The boxplot of the mean *TTC* values within the range of 0 to 2.7 of the three algorithms.

Table 4 Statistical results of the mean *TTC* within the range of 0 to 2.7

| Algorithm | TD3 | HCFS | K-HCFS |
|---|---|---|---|
| Minimum | 0 | 1.0250 | 0.8921 |
| Q1 | 0 | 1.9145 | 1.8500 |
| Medium | 0.9148 | 2.2639 | 2.2122 |
| Q3 | 2.2981 | 2.5075 | 2.4889 |
| Maximum | 2.6999 | 2.6999 | 2.6999 |

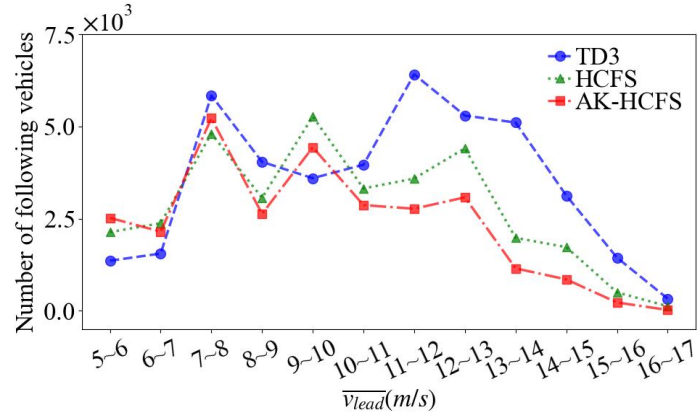

Fig. 5. The number of following vehicles of which *TTC* is within the range of 0 to 2.7 seconds with leading vehicles of different leading speed of the three algorithms. The horizontal axis represents different intervals of mean speed of the leading vehicle and the vertical axis represents the number of the following vehicles with the mean *TTC* within the range of 0 to 2.7.

Fig. 3 shows that AK-HCFS has the least number of following vehicles of which *TTC* between 0~2.7 seconds and it indicates the highest safety. TD3 has the largest number, and most of them are concentrated in the range of 0~0.3 seconds, and the safety is the lowest. AK-HCFS and HCFS successfully reduce the number of vehicles whose mean value of TTC is between 0 ~ 0.3 s, and increase the number of other intervals within 0 ~ 2.7 s. Overall, the safety has been improved. The numerical statistics in Fig. 4 and Table 4 show that both HCFS and AK-HCFS algorithms have greatly improved security. The numerical results of AK-HCFS are similar to those of HCFS. The statistical results of the number of vehicles in Fig. 5 also show that the overall safety performance of AK-HCFS is the best and the number is the least, especially at higher speeds.

(3) Statistical analysis of the absolute value of jerk

Figs. 6, 7, and 8 show the quantitative and numerical statistics of the absolute value of the jerk, and the absolute value of the jerk at different leading vehicle speeds according to the speed of the leading vehicle, respectively. Table 5 shows the minimum, maximum, median, and first quartile Q1 and third quartile Q3 in Fig. 7.

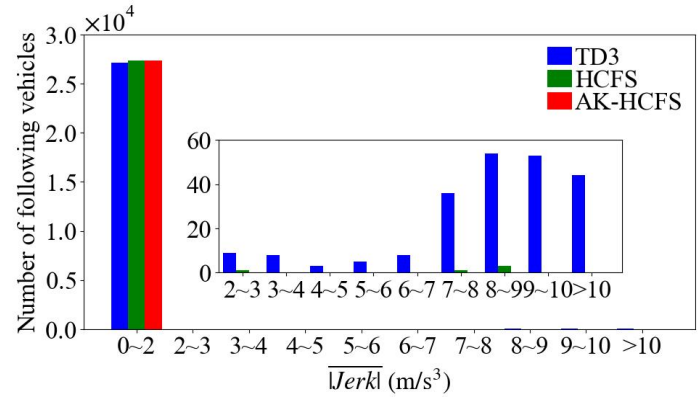

Fig. 6. Statistics of the mean absolute value of jerk of the three algorithms. The horizontal axis represents different intervals of the mean value of the absolute *Jerk* and the vertical axis represents the number of the following vehicles.

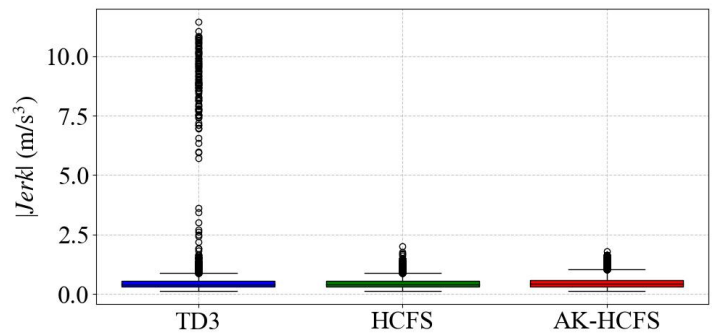

Fig. 7. The boxplot of the mean absolute jerk of the three algorithms.

Table 5 Statistical results of the absolute *Jerk*

| Algorithm | TD3 | HCFS | K-HCFS |
|---|---|---|---|
| Minimum | 0.1294 | 0.1295 | 0.1296 |
| Q1 | 0.3027 | 0.3052 | 0.3079 |
| Medium | 0.3947 | 0.3947 | 0.4134 |
| Q3 | 0.5405 | 0.5363 | 0.5928 |
| Maximum | 0.8972 | 0.8827 | 1.0201 |

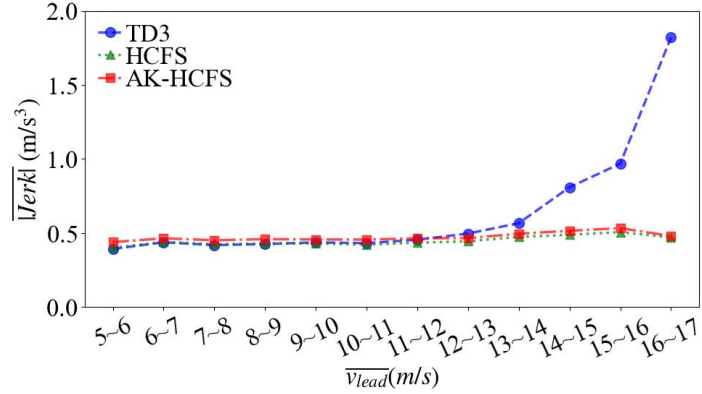

Fig. 8. The mean values of the absolute jerk with leading vehicles of different mean leading speed of the three algorithms. The horizontal axis represents different intervals of mean speed of the leading vehicle and the vertical axis represents the mean values of the absolute jerk.

In automated transportation systems, 2 $m/s^3$ is usually chosen as the comfort requirement (Martinez and Canudas, 2007; Luo et al., 2010). Fig. 6 counts the number of jerk values within 2 $m/s^3$ and above, and it can be seen that the jerk values of AK-HCFS are within this standard, while the TD3 algorithm has a large number of cases that exceed this standard. There is also very small quantity of following vehicles controlled by HCFS that exceed this standard. Fig. 7 and Table 5 show the numerical statistics of the comfort of the three algorithms. The boxplot results of the three algorithms are similar, and the jerk value of AK-HCFS in the statistics is slightly worse than that of the other two algorithms. But it's all within the limit of 2 $m/s^3$. Fig. 8 shows the mean of the absolute values of the jerk at different leading vehicle speeds. Under the control of the TD3 algorithm, at higher speeds, the jerk value increases, which means that the comfort is reduced. However, HCFS and AK-HCFS still maintain stable comfort indicators at high speeds. From the above results, it can be seen that although the AK-HCFS algorithm has an impact on comfort, its influence is controlled within a certain limit.

(4) Statistical analysis of the average speed

Figs. 9, 10, and 11 show the quantitative and numerical statistical results of the mean vehicle speed and the mean speed of the head vehicle at different speeds according to the speed of the head vehicle, respectively. Table 6 shows the minimum, maximum, median, and first quartile Q1 and third quartile Q3 in Fig. 6. It is shown that the number distribution of the mean velocities of the three algorithms is very close, which implies that the AK-HCFS and HCFS algorithms have a very low impact on efficiency.

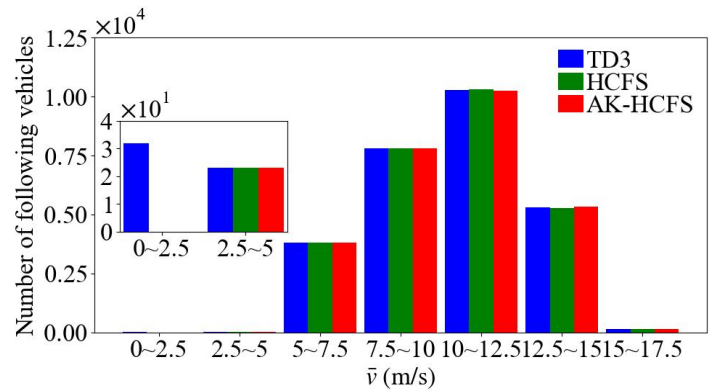

Fig. 9. Statistics of the mean speed of the three algorithms. The horizontal axis represents different intervals of the mean speed and the vertical axis represents the number of the following vehicles.

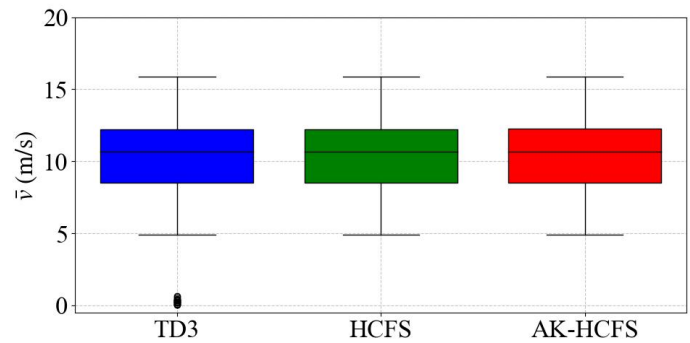

Fig. 10. The boxplot of the mean speed of the three algorithms.

Table 6 Statistical results of the average speed

| Algorithm | TD3 | HCFS | K-HCFS |
|---|---|---|---|
| Minimum | 4.8851 | 4.8851 | 4.8855 |
| Q1 | 8.5053 | 8.5085 | 8.5273 |
| Medium | 10.6468 | 10.6503 | 10.6676 |
| Q3 | 12.2359 | 12.2324 | 12.2495 |

|  | Maximum | 15.8922 | 15.8922 | 15.8922 |

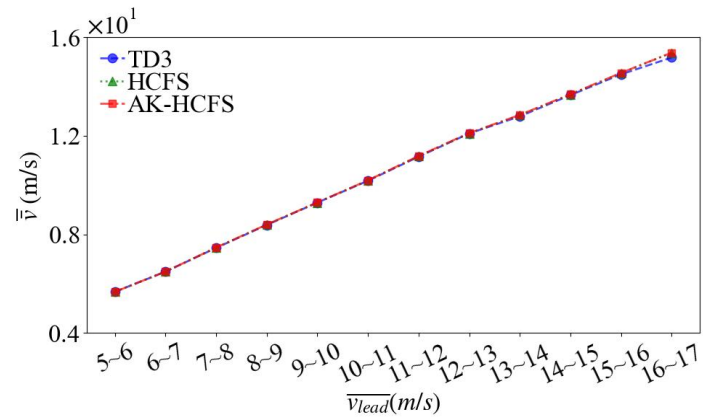

Fig. 11. The mean values of the mean speed with leading vehicles of different mean leading speed of the three algorithms. The horizontal axis represents different intervals of mean speed of the leading vehicle and the vertical axis represents the mean values of the mean jerk of the following vehicles.

Based on the statistical results of the four indicators, it can be seen that the proposed AK-HCFS can significantly improve the overall safety of autonomous driving and avoid collisions with little impact on the overall efficiency. At the same time, the impact on overall comfort is kept to a certain limitation.

B. Representative cases

To compare the three algorithms TD3, HCFS, and AK-HCFS further, we have selected two types of cases for analysis, where Case 1 is the case where the HCFS algorithm has a collision, and Case 2 is a classical case where no collision occurs.

(1) Case 1

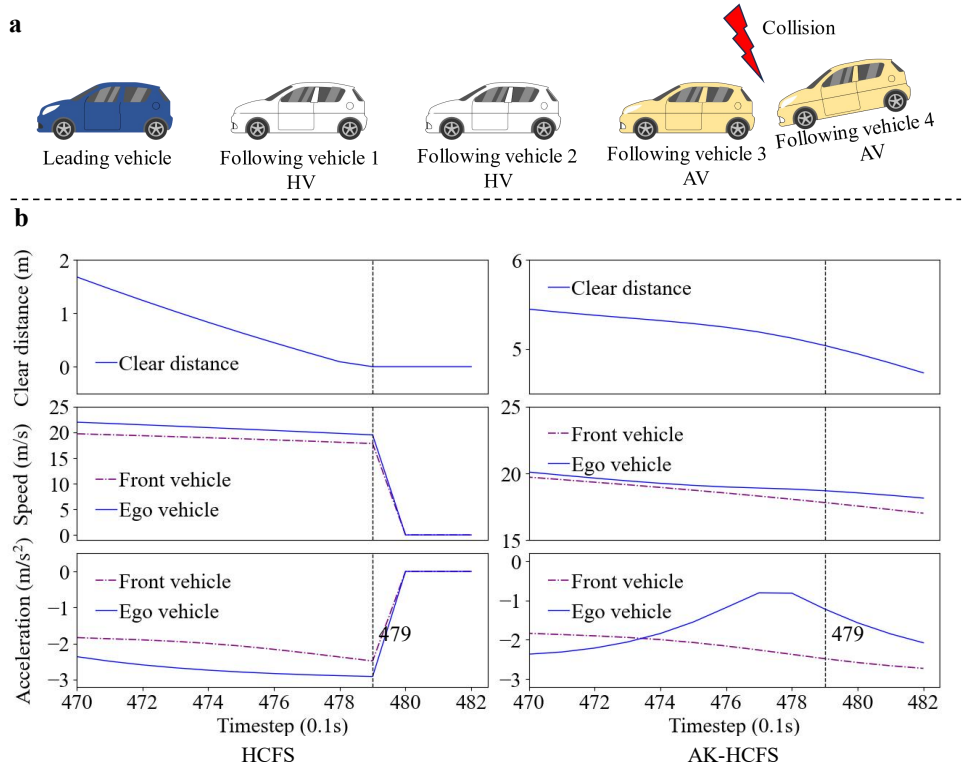

Fig. 12. The clear distance, speed and acceleration of case 1 of HCFS and AK-HCFS around the collision. **a**, The schematic of the collision. The blue vehicle is the leading vehicle and the white ones are the human drive vehicles. The red vehicles are the following vehicles controlled by the autonomous driving algorithms. The collision is between the third and fourth vehicles. **b**, The comparison between HCFS(left) and K-HCFS(right) around the collision. AK-HCFS successfully avoids the collision. In both parts the dashed purple line represents the front vehicle (the third vehicle) and the blue line represents the ego vehicle (the fourth vehicle).

In Case 1, the HCFS algorithm has a collision, and the first two of the four vehicles are autonomous driving vehicles, and the last two are autonomous vehicles. The first row of Fig. 12 shows the scene at the time of the collision, which occurred between the third and fourth following vehicles. Fig. 12 a shows the clear distance between the front and rear vehicles, the speed of the front and rear vehicles, and the acceleration of the front and rear vehicles in the event of a collision with the HCFS algorithm. Fig. 12 b represents the net distance between the front and rear vehicles, the speed of the front and rear vehicles, and the acceleration of the front and rear vehicles of the AK-HCFS algorithm at the same timestep.

It can be seen that the collision occurred at the 479th timestep of this car following event. Before the collision, both the ego and the front vehicle under the control of the HCFS algorithm are driving at a high speed, and both vehicles take deceleration measures. The net distance between the front and rear vehicles is 0.0921 m, the speed of the front vehicle is 18.0649 m/s, and the speed of the ego vehicle is 19.7929 m/s. In this state, the acceleration taken by the vehicle is -2.9147 m/s$^2$. At the time of the collision, the speed of the vehicle in front is 17.8164 m/s. Obviously, in this case, on the 478th timestep, no collision can be avoided by any acceleration within the range of acceleration between the vehicle in front and the vehicle. This is caused by the switching of HCFS on the previous timestep and the ego vehicle on the timestep, which shows that there is a hidden danger in only looking at the return function of the current timestep.

For the ego vehicle and the front vehicle controlled by the AK-HCFS algorithm, the front and rear vehicles maintain a safe distance between the front and rear vehicles at the same timestep, and the front and rear vehicles are at a high speed and are decelerating. In addition, the preceding vehicle speed and acceleration of the two algorithms are similar. Due to the safer distance between the front and rear vehicles, the rear vehicle under the control of the AK-HCFS algorithm also experienced a decrease in the distance from the front vehicle, but there is no collision. This is due to the fact that AK-HCFS adopted a different strategy than HCFS at the previous timestep, and these differences lead to the difference in the final results of the two algorithms.

(2) Case 2

Case 2 is a case where no collision occurs. In this paper, the speed of the leading vehicle is classified according to the mean speed, and a total of 456 groups of leading vehicle data are divided into 13 groups, and the largest number of them is selected (the group has 86 leading vehicle data, accounting for 19% of the total). The mean speed of the first vehicle in this group of data ranges from 12 ~ 13 m/s. In these 86 leading data, the minimum speed and maximum speed of the leading vehicle are further classified. They are divided into 9 groups according to the minimum speed, and the group with the largest number has 27 leading vehicle data. The minimum speed range is 5 ~ 6 m/s. It is divided into 11 groups according to the maximum speed, of which the largest number of groups has 20 leading vehicle data, and the maximum speed range is 22 ~ 23 m/s. The leading vehicle with the largest minimum speed and maximum speed in the group has one set of data, which is the data of the leading vehicle in Case 2.

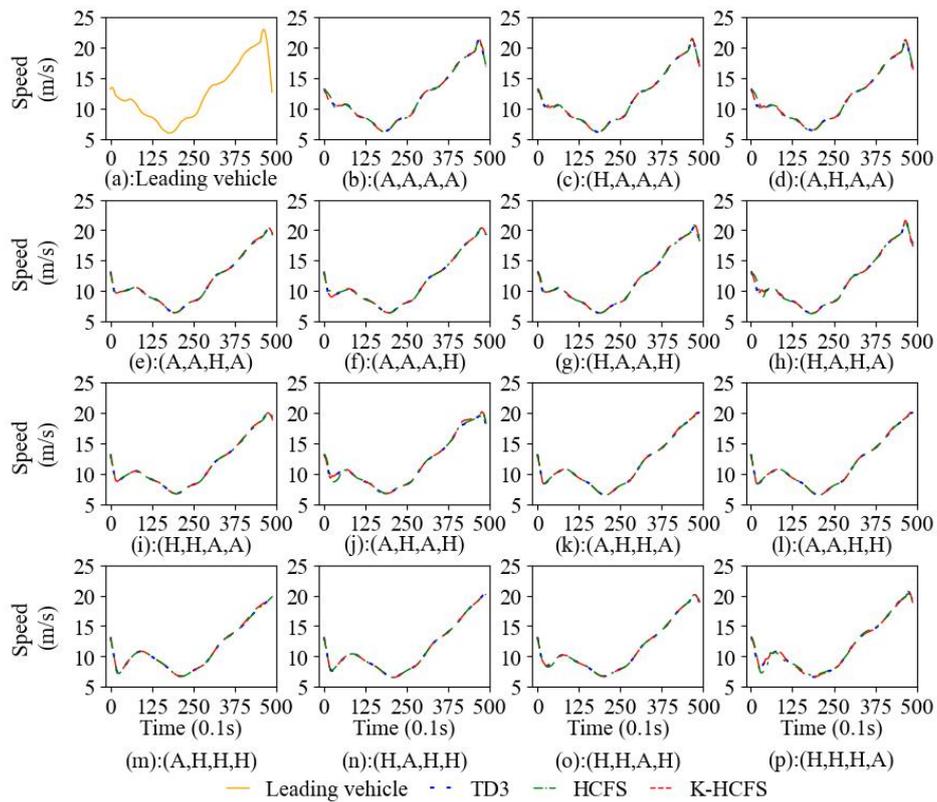

Fig. 13. The speed of the last following vehicle of case 2 under all the mixed traffic scenario of the three algorithms. (a) The speed of the leading vehicle. (b)~(p) The last following vehicle of case 2 under different mixed traffic scenario. A represents autonomous vehicle and H represents human drive vehicle. For example (H, A, A, A) means the first following vehicle is human drive vehicle and the rest are autonomous vehicles. The color yellow, blue, green and red represent the speed of leading vehicle, TD3, HCFS and AK-HCFS respectively.

None of the three algorithms collided in Example 2, and Fig. 13 shows the speed of the last following vehicle of the three algorithms in Example 2 in mixed traffic. Among them, the first picture on the upper right shows the speed curve of the leading vehicle. As one can see that, the speeds of the three algorithms are very close. This confirms that the efficiency of the three algorithms in the statistical results is similar.

## 5. Conclusion

This paper proposes an adaptive Kalman-based hybrid car following strategy for autonomous vehicles, which solves the problem that the traditional hybrid strategy based on fixed coefficients is difficult to adapt to complex environments, resulting in the degradation of car following performance and even unsafe accidents. The strategy adopts the KF to fuse the two signals, CACC and TD3, and realizes the adaptive adjustment of the coefficient $H$, by making multiple-step predictions. In this paper, simulation tests of 4157745 timesteps are conducted under mixed traffic flow scenarios to validate the three algorithms, TD3, HCFS, and AK-HCFS. Statistically, four indicators including collisions in the tests, the mean value of $TTC$, the absolute value of jerk, and the average speed are analyzed. Detailly, two typical cases, with collision and no collision, are analyzed. The analysis of the results shows that the number of collisions is reduced from 58 (TD3) and 1 (HCFS) to 0 (AK-HCFS), and there is a statistically significant reduction in the number of vehicles whose mean value of $TTC$ is less than 2.7 s. Compared with the TD3 and HCFS algorithms, the AK-HCFS algorithm proposed in this paper significantly improves the safety of car following of autonomous vehicles under mixed traffic flow scenarios without compromising on comfort and efficiency.

**Acknowledgement:**

This work is partly supported by the National Natural Science Foundation of China (72288101, 72242102, 71931002) and National-funded Postdoctoral Researcher Program (GZC20233370)., and State Key Laboratory of Automotive Safety and Energy( KFY2208) and Fundamental Research Funds for the Central Universities (2023JBMC043). Finally, the first author would like to acknowledge the State Scholarship Fund provided by the China Scholarship Council that supports her research in Institut Polytechnique de Paris, France.